 \documentclass[final,5p,times,twocolumn]{elsarticle}

\usepackage{amssymb}
\usepackage{lipsum}
\usepackage{subfigure}
\usepackage{multirow}
\usepackage{amsmath}
\usepackage{array}
\begin{document}

\begin{frontmatter}



\title{DIABETIC RETINOPATHY DETECTION USING QUANTUM TRANSFER LEARNING}


% \author[inst1]{ANKUSH JAIN}

% \affiliation[inst1]{organization={University of the Moon},%Department and Organization
%             city={Earth},
\author[1]{ANKUSH JAIN}
\affiliation[1]{ankush.jain@nsut.ac.in, Netaji Subhas University of Technology, New Delhi}
\author[2]{RINAV GUPTA}
\affiliation[2]{rinav.ug20@nsut.ac.in, Netaji Subhas University of Technology, New Delhi}
\author[3]{JAI SINGHAL}
\affiliation[3]{jai.singhal.ug20@nsut.ac.in, Netaji Subhas University of Technology, New Delhi}

% \affil[1]{Department of Mathematics, University X}
% \affil[2]{Department of Biology, University Y}
\begin{abstract}
Diabetic Retinopathy (DR), a prevalent complication in diabetes patients, can lead to vision impairment due to lesions formed on the retina. Detecting DR at an advanced stage often results in irreversible blindness. The traditional process of diagnosing DR through retina fundus images by ophthalmologists is not only time-intensive but also expensive. While classical transfer learning models have been widely adopted for computer-aided detection of DR, their high maintenance costs can hinder their detection efficiency. In contrast, Quantum Transfer Learning offers a more effective solution to this challenge. This approach is notably advantageous because it operates on heuristic principles, making it highly optimized for the task. Our proposed methodology leverages this hybrid quantum transfer learning technique to detect DR. To construct our model, we utilize the APTOS 2019 Blindness Detection dataset, available on Kaggle. We employ the ResNet-18, ResNet34, ResNet50, ResNet101, ResNet152 and Inception V3, pre-trained classical neural networks, for the initial feature extraction. For the classification stage, we use a Variational Quantum Classifier. Our hybrid quantum model has shown remarkable results, achieving an accuracy of 97\% for ResNet-18. This demonstrates that quantum computing, when integrated with quantum machine learning, can perform tasks with a level of power and efficiency unattainable by classical computers alone. By harnessing these advanced technologies, we can significantly improve the detection and diagnosis of Diabetic Retinopathy, potentially saving many from the risk of blindness.
\end{abstract}




\end{frontmatter}

%\tableofcontents


\section{INTRODUCTION}

Diabetes is a condition where there is insufficient insulin, which raises blood glucose levels \cite{taylor2012handbook}. Diabetes has an impact on the kidneys, heart, nerves, and retina \cite{taylor2012handbook} \cite{islam2020diabetes}. 
A diabetic condition known as diabetic retinopathy (DR) results in swelling and blood and fluid leakage from the retina's blood vessels \cite{boyd2020american}. DR was expected to impact around 103.12 million adults worldwide in 2020, with forecasts indicating an increase to 160.50 million by 2045 \cite{teo2021global}. The global prevalence of DR among diabetics is approximately 22.27\%, with varied stages such as proliferative DR (6.96\%) and diabetic macular edema (6.81\%) \cite{teo2021global} \cite{yau2012global}. If DR is diagnosed early, it can be managed using available treatments. Regular eye fundus examination is necessary because DR do not present any symptoms at early stages. Proliferative and non-proliferative DR are the two primary types of DR \cite{qummar2019deep}.

Artificial intelligence (AI)-based algorithms have currently successfully diagnosed several medical conditions, including a variety of retinal illnesses like DR \cite{bellemo2019artificial}. There is a problem with the manual diagnosis of diabetic retinal disease (DR) based on retinal pictures; it is a laborious procedure that lacks medical specialists. Therefore, it is ideal to create an automated DR detection system that could work with medical professionals' help in order to overcome the current obstacles \cite{gangwar2021diabetic}. Many efforts have been made to automate the classification of DR images using deep learning to help ophthalmologists identify disease early on. CNN is one of the most popular deep learning algorithms since it has proven to be effective and successful in image analysis \cite{alyoubi2021diabetic}. CNNs are DL-based, futuristic models that have led to numerous advances in automated object recognition and classification. These deep learning-based methods have the potential to extract useful characteristics for precise image classification. As a result, multi-path CNN was created to extract DR features from retinal pictures, which could then be used in machine learning to carry out DR classification \cite{gayathri2021diabetic}.

As the first small-scale quantum computers have advanced, quantum DL techniques are currently receiving a lot of attention. Scholars have developed a number of categorization models based on different quantum parametric circuits, in which conventional data is identified as unique qubits \cite{mathur2021medical}. In many applications, quantum computers have shown to be more reliable than conventional computers, particularly when sampling complex probabilities. Therefore, it is common to inquire whether this hierarchy makes use of learning models. A resounding yes is widely taken to be true, yet there appears to be little study being done in this area to produce unambiguous representations of how to achieve this quantum merit. According to some claims, quantum-based techniques on classical computers have the advantage of exposing effective solutions, making picture classification a non-exception \cite{mangini2021quantum}.

While several efforts have been made to automatically classify DR images using DL in order to help ophthalmologists identify the disease at an early stage, the majority of these efforts have focused just on DR detection rather than the identification of multiple DR phases. Additionally, there aren't many challenges in identifying and locating the different forms of DR lesions, which is useful in practice since opthalmologists can evaluate the severity of DR and track its evolution based on the appearance of the lesion. Due to these factors, the current study suggests using quantum-based CNN for fully automated screening in order to simultaneously localize all DR lesion types and detect the five phases of DR. These goals are outlined below. The suggested effort would help opthalmologists mimic the DR diagnostic technique that locates DR lesions, identifies their type, and determines the true phase of DR. The following is a list of this study's primary contributions:
\begin{itemize}
    \item Resizing the photos is done during pre-processing to provide them flexibility for additional processing;

\item To carry out DR classification using the suggested quantum-based Deep CNN, which enhances the accuracy of the suggested system by utilizing an optimized multiple-qubit gate quantum neural network;
\item The effectiveness of the suggested model is disclosed in order to assess its efficacy using common performance metrics including recall, accuracy, specificity, precision, and f1-score.
\end{itemize}

 \begin{figure}[ht]
	\centering 
	\includegraphics[width=0.35\textwidth]{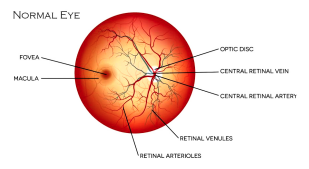}	
	\caption{Normal Eye Retina} 
    \end{figure}
    \begin{figure}[ht]
        \centering 
	\includegraphics[width=0.35\textwidth]{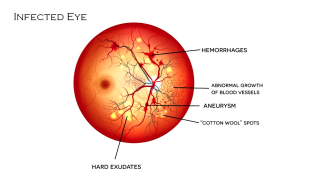}	
	\caption{Infected Eye Retina}
    \end{figure}

The structure of the paper is as follows: The literature study and transfer learning overview are covered in Section 2. Section 3 follows, in which the suggested system is explained using the appropriate flow and mathematical representations. Results obtained following the implementation of the suggested system are contained in Section 4. Finally, Section 5 provides a summary of the research findings together with suggestions for the future.

\section{LITERATURE REVIEW}
%%\label{}

Transfer learning is a predictive modeling method in which a model established for one problem is reused or changed for another, increasing training speed and performance. This method is especially useful for training deep neural networks with minimal data sets. Transfer learning (TL) with convolutional neural networks (CNNs) is important in medical image analysis because it uses knowledge from previous jobs to improve performance on new ones. It addresses issues like data scarcity and the requirement for considerable hardware resources. Traditional feature extraction approaches in medical image analysis have given way to deep learning (DL) algorithms like CNNs, which require big datasets for optimal training. Domain adaptation and unsupervised learning, for example, assist circumvent this data barrier by adapting information from source tasks to target tasks \cite{kim2022transfer}. The two basic ways to utilizing CNN models in TL are feature extraction and fine-tuning, with each giving distinct advantages in terms of computing efficiency and adaptability \cite{kim2022transfer}.

\begin{figure}[!t]
    \centering
    \includegraphics[width=0.4\textwidth]{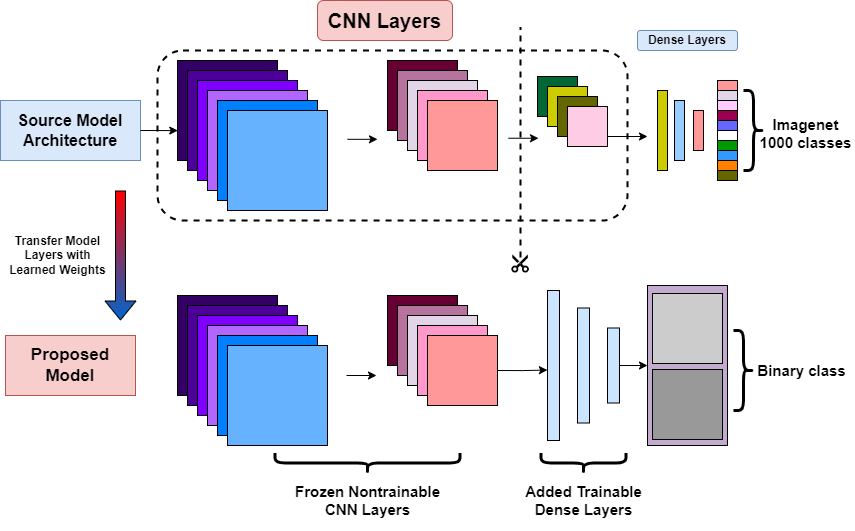}
    \caption{
    {Architecture of Transfer Learning Model.}
    }
\end{figure}

Quantum machine learning extends this paradigm, particularly in hybrid neural networks that combine conventional and quantum elements. A quantum layer is frequently added to a classically pre-trained network in this case. To categorize the retrieved characteristics, this method employs a quantum classifier, such as a variational quantum classifier.

\begin{table*}
\centering
\renewcommand{\arraystretch}{1.35}
\begin{tabular}{|p{0.15\textwidth}|p{0.25\textwidth}|p{0.55\textwidth}|}
\hline
\textbf{Author} & \textbf{Adopted models} & \textbf{Experimental out-comes} \\
\hline
Mir et al. \cite{mir2022diabetic} & Inception-V3 and Variational Quantum Classifier & Accuracy of 93\%–96\% on the quantum hybrid model and 85\% accuracy rate on the classical model \\
Mohammadian et al. \cite{mohammadian2017comparative} & Inception-V3 and Xception pre-trained models & Fine-tuned the Inception-V3 and Xception pre-trained models to classify the DR data set into two classes. Accuracy score of 87.12\% and 74.49\% achieved. \\
Wan et al. \cite{wan2018deep} & Pre-trained models AlexNet, VggNet-s, VggNet-16, VggNet-19, GoogleNet and RestNet & Implemented transfer learning and hyperparameter tuning on the pre-trained models. Highest accuracy score was that of VggNet-s model, which reached 95.68\%. \\
Dutta et al. \cite{dutta2018classification} & Shallow feed forward neural network, deep neural network and VggNet-16 model & On a test data set of 300 images, the shallow neural network scored an accuracy of 42\%, and the deep neural network scored 86.3\% while the VggNet-16 scored 78.3\% accuracy. \\
Gangwar and Ravi \cite{gangwar2021diabetic} & Pretrained Inception-Resnet-V2 and some custom block of CNN layers & The model performed better on APTOS 2019 dataset with accuracy of 82.18\%. \\
T. Shahwar et al. \cite{shahwar2022automated} & Pre-trained model ResNet 34 and vector to quantum variational circuit (QVC) & A hybrid classical-quantum model was proposed achieving a precision of 99.1\%. \\
Gondal et al. \cite{gondal2017weakly} & A CNN model & The performance of the CNN model is evaluated based on binary classification resulting in sensitivity 93.6\% and specificity 97.6\% on DiaretDB1. \\
Wang et al. \cite{wang2017zoom} & An Inception Model & High AUC on a normal and referable DR task 0.978 and 0.960 respectively and specificity is 0.5. \\
Chanrakumar and Kathirvel \cite{chandrakumar2016classifying} & CNN model with a dropout regularization technique & The accuracy achieved by their model is 94\%. They manually performed augmentation and preprocessing steps by using an image editing tool. \\
Memon et al. \cite{memon2017diabetic} & CNN Architecture & The overall kappa score accuracy is 0.74, for the validation purpose, 10\% of the images were used. \\
Garcia et al. \cite{garcia2017detection} & CNN (Alexnet, VGGnet16, etc.) & Achieved the best results on VGG16 and achieved 93.65\% specificity, 54.47\% sensitivity, and 83.68\% accuracy. \\
Kumar et al. \cite{kumar2020automated} & Neural network & The results have exhibited the advancement of the considered methodology with accuracy as the performance measure. \\
Thomas et al. \cite{grasbeck2016fundus} & Fundus images based screening & The implementation results demonstrate that just 6 of the underlying DR acknowledgment was ineffectual, accordingly uncovering its advancement in treating DR in youngsters. \\
Gupta et al. \cite{gupta2017local} & Random forest approach & The introduced methodology was assessed by considering both clinic and public datasets with measures sensitivity and specificity. \\
Georgios et al. \cite{leontidis2017new} & Random forest approach & The methodology concentrates on examining the high and low points of DR for perceiving the vascular variations. Likewise, Retinal fundus pictures were analyzed. \\
Manuel et al. \cite{gegundez2017tool} & CAD & The investigation results have offered a superior degree of sensitivity, which was seen as the one accomplished by the specialists. \\
Welikala et al. \cite{welikala2015automated} & Multi layered perceptron network & An algorithm based on an SVM classifier which utilizes a Gabor filtering approach. \\
Shanthi and Sabeenian \cite{shanthi2019modified} & Convolutional neural network & The precision of the executed model was demonstrated from the implemented results. \\
Zago et al. \cite{zago2020diabetic} & Convolutional neural network & The computations related to the received recognizing strategy were implemented and the upgraded results were achieved for sensitivity and precision. \\
\hline
\end{tabular}
\caption{Survey Papers on DR Detection}
\label{survey papers}
\end{table*}

Several DR detection research have made use of traditional machine learning approaches. For instance, Mansour \cite{mansour2018deep} deployed deep CNNs with transfer learning for DR diagnosis. Mohammadian \cite{mohammadian2017comparative} applied fine-tuning on pre-trained models like Inception-V3 and Xception for DR classification, achieving notable accuracy. On the Kaggle dataset, Wan et al. \cite{wan2018deep} used transfer learning with models such as AlexNet and VggNet, with Vgg Net-s demonstrating the highest accuracy. Dutta et al. \cite{dutta2018classification} used fundus pictures to train several models, including VggNet-16, with varying degrees of accuracy.

These papers emphasize the widespread usage of transfer learning in DR detection, with a focus on model performance rather than dataset quality improvement. Transfer learning model accuracy is largely dependent on dataset properties. Google's quantum supremacy and the creation of new quantum algorithms like parameterized quantum circuits (PQCs) and Quantum Neural Networks (QNNs) mimic the evolution from machine learning to deep learning, unlocking tremendous computing possibilities.

Gangwar et al. \cite{gangwar2021diabetic} used a model that used the pretrained Inception-Resnet-V2 with additional custom CNN layers to detect blindness in the Messidor-1 diabetic retinopathy dataset and APTOS 2019 blindness detection (Kaggle dataset). The inception blocks processed photos of 229x229 pixels using filters of sizes 3x3, 5x5, 7x7, and 9x9. On the APTOS 2019 dataset, this model achieved an accuracy of 82.18%.

T. Shahwar et al. \cite{shahwar2022automated} used 6400 tagged MRI scans to construct a hybrid classical-quantum model for Alzheimer's detection in the world of quantum computing. A pre-trained ResNet 34 was used for feature extraction, and a quantum variational circuit (QVC) was used to reduce a 512-feature vector to four features. The model had a training accuracy of 99.1%.
Using the APTOS 2019 dataset and Inception V3 for feature extraction, a classical-quantum transfer learning model in ophthalmology \cite{mir2022diabetic} revealed significant improvements in 2022. This model achieved accuracies of 93-96\% when tested on several quantum simulation systems.

Sim et al. \cite{sim2019expressibility} investigated the properties of variational quantum circuits, as did \cite{hubregtsen2021evaluation}. They investigated the expressibility and entanglement capacity of these circuits, discovering a substantial association between circuit expressiveness and classification accuracy and a lower correlation between entanglement capacity and classification accuracy. This discovery influenced the design of quantum circuits for classification tasks, particularly in hybrid models where pre-trained classical models are utilized for feature extraction and variational quantum circuits are used for classification.

Gondal et al. \cite{gondal2017weakly} created a CNN model for referable DR that was trained on the Kaggle dataset and tested on DiaretDB1. Normal and mild DR were classed as non-referable, whereas the remaining stages were classified as referable, yielding 93.6\% sensitivity and 97.6\% specificity on DiaretDB1. Wang et al. \cite{wang2017zoom} proposed a novel architecture for categorizing DR pictures, combining main, attention, and crop networks to achieve excellent AUC scores.

Quelle et al. \cite{quellec2017deep} created three CNN models for binary classification, with a focus on detecting DR lesions, using both the Kaggle and DiaretDb1 datasets. \cite{chandrakumar2016classifying} constructed a CNN model with dropout regularization on the Kaggle dataset and tested it on the DRIVE and STARE datasets, reaching 94\% accuracy. \cite{memon2017diabetic} used CNN on the Kaggle dataset as well, using preprocessing techniques such as nonlocal mean denoising, and found a kappa score accuracy of 0.74. \cite{pratt2016convolutional} presented a CNN architecture for distinguishing the five stages of DR, but had difficulty reliably diagnosing the moderate stage.

Yang et al. \cite{yang2017lesion} investigated the use of a Deep Convolutional Neural Network (DCNN) for two stages of DR, using a local-global network strategy for lesion highlighting and grading. The performance of several CNN models on the Kaggle dataset was examined in research \cite{dutta2018classification}. Garcia et al. \cite{garcia2017detection} used right and left eye pictures individually, employing CNN models such as AlexNet and VGGnet16, with VGG16 producing the best results without completely connected layers. Dutta et al. 
 \cite{dutta2018classification} used the Kaggle dataset with three deep learning models: FNN, DNN, and CNN, picking 2000 photos from the dataset for training and getting the maximum training accuracy with DNN.

% A random equation, the Toomre stability criterion:

% \begin{equation}
%     Q = \frac{\sigma_v \times \kappa}{\pi \times G \times \Sigma}
% \end{equation}

\section{PROPOSED METHODOLOGY}
%%\label{}
Quantum computing, with its unique capabilities using qubits, offers exponential processing power compared to classical computing, particularly valuable for complex problem-solving across various disciplines, including healthcare. The integration of quantum transfer learning enhances quantum algorithm development by leveraging pre-trained models, reducing resource demands, and accelerating progress in chemistry, optimization, and machine learning within the quantum computing landscape. The study delves into classical model foundations, quantum circuit integration with classical networks, and the application of hybrid models in tasks like diabetic retinopathy detection, illustrating the potential transformative impact of quantum technologies on medical diagnostics and beyond.

\subsection{Quantum Computing vs. Classical Computing}

Quantum computing is a major leap forward from classical computing, employing qubits that may exist in several states at the same time, allowing for exponential improvements in processing capacity. This quantum advantage is especially useful in addressing complicated problems more efficiently than classical computers, and it has potential applications in a variety of disciplines, including healthcare. Quantum computers, for example, may simulate molecular biological or chemical systems, which is useful in drug development and medical diagnostics. However, hybrid algorithms mixing quantum and classical computing approaches are developing as a feasible solution for practical applications in industrial and medical areas due to their current small size and error-prone character.

\begin{figure}[ht]
    \centering
    \includegraphics[width=0.45\textwidth]{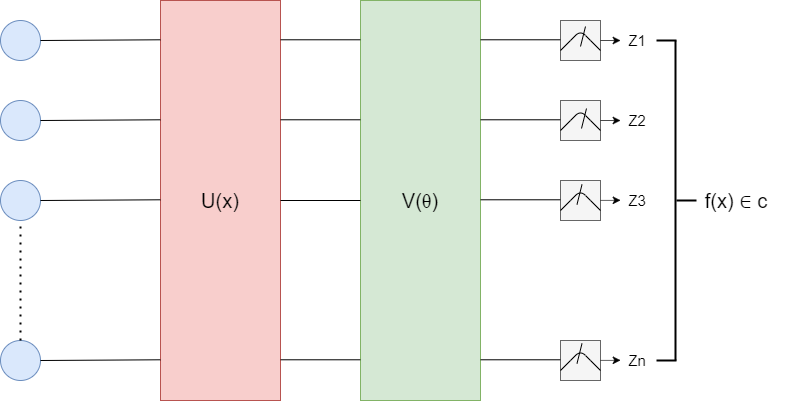}
    \caption{
    {Variational Quantum Classifier with embedding layers $U(x)$ and variational circuit $V(\theta)$ and final measurements in classical output $f(x) \in \mathbb{C}$.}
    }
\end{figure}

 We presented three essential components (embedding layers U(x), variational circuit V($\theta$), and final measurements) on which the Variational Quantum Classifier (VQC) is developed in Figure 4. The variational circuit is an essential component of VQC. Figure 5 depicts a variational circuit for a single qubit operation.
\begin{figure}[ht]
    \centering
    \includegraphics[width=0.45\textwidth]{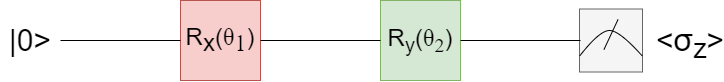}
    \caption{Simple case of one Qubit}
    
    \label{fig:qubit}
    
\end{figure}

\subsection{Quantum Transfer Learning}

Quantum transfer learning is a novel approach that leverages pre-trained quantum models to enhance the training and performance of new quantum tasks. Similar to classical transfer learning, where knowledge gained from one task is transferred to another related task, quantum transfer learning adapts existing quantum models by fine-tuning their parameters or reusing learned features for new applications. This technique accelerates the development of quantum algorithms, reduces the need for extensive quantum resources, and holds promise for advancing various fields, including chemistry, optimization, and machine learning, within the burgeoning realm of quantum computing.
\subsubsection{Classical Model Foundation} 
The classical model foundation for this approach involves the selection of a pre-trained network such as ResNet18, renowned for its strong performance in deep learning tasks related to image processing. This choice is based on the network's established capabilities and effectiveness in handling complex visual data. The model's initial training on a vast dataset like ImageNet ensures that it has acquired a sophisticated ability to recognize and extract detailed features from images. By leveraging this pre-trained architecture and its feature extraction mechanism, we can efficiently utilize learned representations to address specific image processing tasks with enhanced accuracy and efficiency. This foundational framework provides a solid starting point for further optimization and customization tailored to specific application domains.

\subsubsection{Quantum Circuit Integration}
In the integration of the quantum circuit with the classical model, a sophisticated hybrid system is formulated. This hybrid model combines the strengths of classical deep learning with the unique capabilities of quantum computing. By leveraging quantum properties such as superposition and entanglement, the system processes and analyzes the intricate features extracted by the classical model at a quantum level. The quantum circuit serves as a complementary component, enhancing the overall computational power and potential for tackling complex image processing tasks. This integration marks a significant advancement in harnessing the benefits of both classical and quantum computing paradigms, paving the way for innovative solutions in image analysis and beyond.

\subsection{Classical Pre-Processing} 

In the initial stages of retinal image processing for diabetic retinopathy (DR) detection, a critical step involves utilizing a pre-trained model like ResNet18's architecture to preprocess these images. This process is pivotal for extracting essential information needed for subsequent analysis. The primary objective of this phase is to distill a set of 512 high-level features from the retinal images. These features are carefully selected as they hold significant relevance and granularity for the subsequent quantum processing that follows. By leveraging the pre-trained model, the system effectively abstracts and isolates these pertinent features, optimizing the data for further analysis and classification in the subsequent stages of the DR detection pipeline. This streamlined approach ensures that the extracted features are both comprehensive and focused, setting the stage for precise and effective utilization in the subsequent phases of the diagnostic process.

\begin{figure}[ht]
    \centering
    \includegraphics[width=0.4\textwidth]{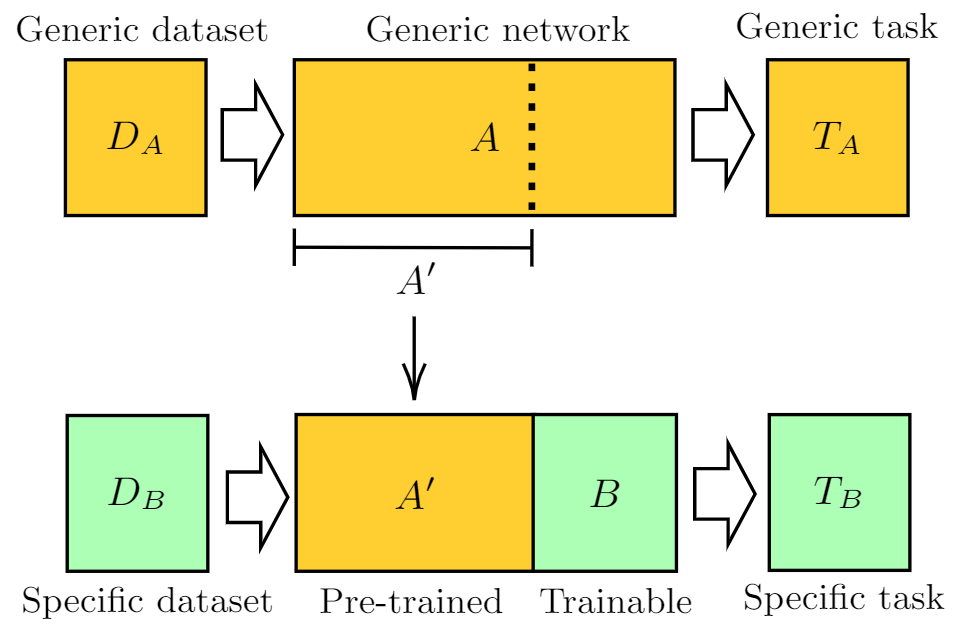}
    \caption{General Transfer Learing}
    \label{transfer learning}
\end{figure}

As sketched in Figure 6, one can give the following general definition of the transfer learning method:
\begin{enumerate}
    \item Take a network A that has been pre-trained on a dataset $D_A$. and for a given task $T_A$.
    \item Remove some of the final layers. In this way, the resulting truncated network A' can be used as a feature extractor.
    \item Connect a new trainable network B at the end of the pre-trained network A'.
    \item Keep the weights of A' constant, and train the final block B with a new dataset $D_B$. and/or for a new task of interest $T_B$.
\end{enumerate}

\subsection{Quantum Circuit for Classification}

    Reshaping Weights: The flat array of quantum weights q\_weights\_flat is reshaped to fit the structure of the variational layers, with dimensions (q\_depth, n\_qubits). This allows each layer of the circuit to have its own set of rotation angles for the qubits.
    
    Embedding Layer: The circuit initializes all qubits in a balanced superposition of up and down states using the Hadamard layer (H\_layer(n\_qubits)). Then, it rotates each qubit around the y-axis by an angle corresponding to the input features (RY\_layer(q\_input\_features)). This process encodes the classical input data into the quantum state of the system.
    This layer can use the following Gates:
    
    Hadamard Gate: The Hadamard gate is one of the fundamental quantum gates, and its role is to create superpositions. When applied to a qubit initially in a basis state ($|$0⟩ or $|$1⟩), the Hadamard gate transforms it into a state that is an equal superposition of the $|$0⟩ and $|$1⟩ states.
    \begin{equation}
        \text{H} = \frac{1}{\sqrt{2}}  \begin{bmatrix} 1 & 1 \\ 1 & -1 \end{bmatrix} 
    \end{equation}
    \textit{where: H represents Hadamard Gate.}\\

    S (Phase) Gate and S† (Dagger) Gate: By applying an S gate followed by a Hadamard, or vice versa, you can introduce a phase shift into the superposition, which can be useful for certain algorithms.

    \begin{equation}
    S = \begin{bmatrix} 1 & 0 \\ 0 & i \end{bmatrix}
    \end{equation}
    \textit{where: The $S$ gate applies a $\pi/2$ phase shift to the $|1\rangle$ state (or equivalently, a $90^\circ$ rotation around the Z-axis on the Bloch sphere), leaving the $|0\rangle$ state unchanged.}
    
    \begin{equation}
    S^\dagger = \begin{bmatrix} 1 & 0 \\ 0 & -i \end{bmatrix}
    \end{equation}
    \textit{where: The $S^\dagger$ gate (conjugate transpose of $S$) applies a $-\pi/2$ phase shift to the $|1\rangle$ state (or a $-90^\circ$ rotation around the Z-axis on the Bloch sphere), leaving the $|0\rangle$ state unchanged.}\\
    
    RX Gate: The RX gate is a rotation around the x-axis of the Bloch sphere. Like the Hadamard gate, it can create superpositions, but with the ability to adjust the ratio of $|$0⟩ and $|$1⟩ in the superposition by changing the rotation angle.

    \begin{equation}
    \text{Rx}(\theta) = \begin{bmatrix} \cos\left(\frac{\theta}{2}\right) & -i\sin\left(\frac{\theta}{2}\right) \\ -i\sin\left(\frac{\theta}{2}\right) & \cos\left(\frac{\theta}{2}\right) \end{bmatrix}
    \end{equation}
    \textit{where:
    \( \theta \) is the rotation angle in radians.
    \( \cos\left(\frac{\theta}{2}\right) \) represents the cosine of half the rotation angle.
    \( \sin\left(\frac{\theta}{2}\right) \) represents the sine of half the rotation angle.
    \( -i \) denotes the imaginary unit.}\\

    RY Gate: Similar to the RX gate, but it rotates around the y-axis. This gate is particularly versatile for creating arbitrary superpositions based on the rotation angle.

    \begin{equation}
    \text{Ry}(\theta) = \begin{bmatrix} \cos\left(\frac{\theta}{2}\right) & -\sin\left(\frac{\theta}{2}\right) \\ \sin\left(\frac{\theta}{2}\right) & \cos\left(\frac{\theta}{2}\right) \end{bmatrix}
    \end{equation}
    \textit{where:
    \( \theta \) is the rotation angle in radians.
    \( \cos\left(\frac{\theta}{2}\right) \) represents the cosine of half the rotation angle.
    \( \sin\left(\frac{\theta}{2}\right) \) represents the sine of half the rotation angle.}\\

    Variational Layers: The circuit iteratively applies a sequence of variational layers, each consisting of an entangling layer followed by a rotation layer. The entangling layer (entangling\_layer(n\_qubits)) applies a predefined pattern of CNOT gates to generate quantum entanglement between qubits. The rotation layer (RY\_layer(q\_weights[k]) then applies rotations around the y-axis to each qubit, with the rotation angles determined by the current set of trainable weights. This sequence is repeated q\_depth times, where q\_depth represents the depth of the quantum circuit.
    This layer can use the following Gates:
    
    CNOT Gate: The CNOT gate is a two-qubit gate that flips the state of the second qubit (target) if the first qubit (control) is in the state $|$1⟩.

    \begin{equation}
    \text{CNOT} = \begin{bmatrix} 1 & 0 & 0 & 0 \\ 0 & 1 & 0 & 0 \\ 0 & 0 & 0 & 1 \\ 0 & 0 & 1 & 0 \end{bmatrix}
    \end{equation}
    \textit{where:
    The rows and columns of the matrix correspond to the basis states \( |00\rangle, |01\rangle, |10\rangle, |11\rangle \) respectively.
    The CNOT gate flips the state of the target qubit (second qubit) if the control qubit (first qubit) is \( |1\rangle \).}\\

    CZ (Controlled-Z) Gate: This gate applies a phase shift only when both qubits are in the $|$1⟩ state. It can be used to create a phase-entangled state and is useful in algorithms that require phase manipulation.

    \begin{equation}
    \text{CZ} = \begin{bmatrix} 
    1 & 0 & 0 & 0 \\ 
    0 & 1 & 0 & 0 \\ 
    0 & 0 & 1 & 0 \\ 
    0 & 0 & 0 & -1 
    \end{bmatrix}
    \end{equation}
    \textit{where:
    The rows and columns of the matrix correspond to the basis states \( |00\rangle, |01\rangle, |10\rangle, |11\rangle \) respectively. The CZ gate applies a phase factor of \( -1 \) to the state \( |11\rangle \), leaving the other basis states unchanged.}\\
    
    SWAP Gate: The SWAP gate exchanges the states of two qubits. It can be used to entangle qubits and is especially useful in algorithms requiring qubit reordering or in architectures with limited qubit connectivity.

    \begin{equation}
    \text{SWAP} = \begin{bmatrix} 
    1 & 0 & 0 & 0 \\ 
    0 & 0 & 1 & 0 \\ 
    0 & 1 & 0 & 0 \\ 
    0 & 0 & 0 & 1 
    \end{bmatrix}
    \end{equation}
    \textit{where:
    The rows and columns of the matrix correspond to the basis states \( |00\rangle, |01\rangle, |10\rangle, |11\rangle \) respectively. The SWAP gate swaps the states of the qubits represented by the second and third basis states \( |01\rangle \) and \( |10\rangle \), while leaving \( |00\rangle \) and \( |11\rangle \) unchanged.}\\
    
    Controlled RX, RY, RZ Gates: These are controlled versions of the rotation gates (RX, RY, RZ) that apply a rotation around the respective axis only when the control qubit is in the $|$1⟩ state. They offer more control over the entanglement process and can be used to create various entangled states.

    \begin{equation}
        \text{CRx}(\theta) = \begin{bmatrix} 
        1 & 0 & 0 & 0 \\ 
        0 & \cos\left(\frac{\theta}{2}\right) & 0 & -i\sin\left(\frac{\theta}{2}\right) \\ 
        0 & 0 & 1 & 0 \\ 
        0 & -i\sin\left(\frac{\theta}{2}\right) & 0 & \cos\left(\frac{\theta}{2}\right) 
        \end{bmatrix}
    \end{equation}\\
    \begin{equation}
        \text{CRy}(\theta) = \begin{bmatrix} 
        1 & 0 & 0 & 0 \\ 
        0 & \cos\left(\frac{\theta}{2}\right) & 0 & -\sin\left(\frac{\theta}{2}\right) \\ 
        0 & 0 & 1 & 0 \\ 
        0 & \sin\left(\frac{\theta}{2}\right) & 0 & \cos\left(\frac{\theta}{2}\right) 
        \end{bmatrix}
    \end{equation}\\
    \begin{equation}
        \text{CRz}(\theta) = \begin{bmatrix} 
        1 & 0 & 0 & 0 \\ 
        0 & e^{-i\frac{\theta}{2}} & 0 & 0 \\ 
        0 & 0 & 1 & 0 \\ 
        0 & 0 & 0 & e^{i\frac{\theta}{2}} 
        \end{bmatrix}
    \end{equation}

    \textit{where:\(\theta\) is the rotation angle in radians. The controlled gates perform a rotation on the target qubit conditioned on the state of the control qubit. The matrices are represented in the computational basis states \( |00\rangle, |01\rangle, |10\rangle, |11\rangle \).}\\

    Measurement Layer: Finally, the circuit measures the expectation value of the Pauli-Z operator on each qubit (qml.expval(qml.PauliZ(position))). These measurements produce a classical output vector where each element corresponds to the expectation value measured on one of the qubits. This vector can represent the output of the quantum circuit for further processing or decision-making in a hybrid quantum-classical algorithm.

\subsection{Hybrid Model Structure}

In dual network integration, a Classical Network (A') extracts features from input data using a truncated pre-trained network like ResNet18. These features are then fed into a Quantum Network (B) for classification using a quantum circuit. Training and optimization of the Quantum Network (B) involve adjusting parameters through techniques like gradient descent to improve classification accuracy based on the extracted features from the Classical Network (A').

\subsubsection{Dual Network Integration}
In the dual network integration approach described, there are two distinct components working in tandem: the Classical Network (referred to as A') and the Quantum Network (referred to as B).

The Classical Network, based on a truncated pre-trained model like ResNet18's architecture, is primarily responsible for feature extraction from input data. Unlike a complete model like ResNet18, which would typically include classification layers, this truncated version focuses solely on extracting meaningful features from the data without making final classifications. This classical feature extraction phase aims to distill relevant information from the input, preparing it for further processing.

Once feature extraction is completed by the Classical Network, the extracted features are then handed over to the Quantum Network for the classification task. The Quantum Network operates using a quantum circuit designed specifically for classification purposes. This circuit leverages the features provided by the Classical Network to perform the final classification in a quantum computing framework.

\subsubsection{Training \& Optimization}
In the training and optimization phase of this dual network integration model, the focus is on refining the Quantum Network (referred to as B) to minimize classification errors efficiently.

Firstly, during training, the primary adjustment targets the parameters of the quantum circuit within the Quantum Network. Unlike the Classical Network (A'), which is a pre-trained model like ResNet18 which focused on feature extraction, the training process specifically modifies and optimizes the parameters of the quantum circuit. By adjusting these parameters, the goal is to enhance the quantum circuit's ability to accurately classify the features extracted by the Classical Network.

To achieve this optimization, various techniques are employed, including gradient descent and its variants. These optimization methods are instrumental in iteratively adjusting the parameters of the quantum circuit to minimize the loss function associated with classification errors. Gradient descent, in particular, allows for efficient updates of the quantum circuit's parameters by computing the gradient of the loss function with respect to these parameters. Through this iterative process, the quantum circuit's parameters are tuned to enhance classification performance and reduce errors.

\subsection{Classical - to - Quantum Transfer Learning}

We focus on the CQ transfer learning scheme and we give a specific example.
\begin{enumerate}
    \item As pre-trained network A we use ResNet18, a deep residual neural network introduced by Microsoft \cite{he2016deep} which is pre-trained on the ImageNet dataset. Apart from ResNet18, we perform a comparative study across ResNet34, ResNet50, ResNet101, ResNet152 and Inception V3.
    \item After removing its final layer we obtain A', a pre-processing block which maps any input high-resolution image into 512 abstract features
    \item Such features are classified by a 4-qubit “dressed quantum circuit” B, i.e., a variational quantum circuit sandwiched between two classical layers
    \item The hybrid model is trained, keeping A' constant, on the Kaggle Dataset containing fundus images of the retina.
\end{enumerate}
A graphical representation of the full data processing pipeline is given in the figure below.
\begin{figure}[ht]
    \centering
    \includegraphics[width=0.4\textwidth]{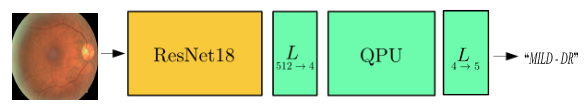}
    \caption{Classic-to-Quantum Transfer Learning}
    \label{CQ}
\end{figure}

\subsection{Application to DR Detection}
The integrated model for diabetic retinopathy (DR) detection specializes in stage-wise classification, crucial for accurate diagnosis. It incorporates a quantum circuit trained to recognize patterns indicating different DR stages, learning from labeled datasets to classify features accurately.

This approach enhances diagnostic capabilities, enabling early detection and intervention for effective patient care. By leveraging the quantum circuit's ability to discern subtle features associated with DR stages, the model supports clinicians in making informed decisions for optimal management and treatment strategies.

\subsection{Algorithm Steps}
This algorithm outlines a systematic approach for diabetic retinopathy (DR) detection and classification. The steps include
\begin{enumerate}
    \item Data Pre-processing: Detailed feature extraction from DR images using the classical network.
    \item Quantum Classification: In-depth classification of features by the quantum circuit.
    \item Training and Validation: Iterative optimization of the quantum circuit with a focus on reducing classification errors.
    \item Evaluation Metrics: Utilizing accuracy, precision, recall, and other metrics to assess the model's effectiveness.
\end{enumerate}

\subsection{Conclusion}
The synergy between classical and quantum computing represents a significant advancement in the field of medical imaging, potentially surpassing the capabilities of either computing approach in isolation. This combined methodology harnesses the strengths of classical deep learning for robust feature extraction and quantum computing for intricate classification tasks, creating a powerful tool with enhanced capabilities for medical diagnostics.

This innovative approach holds promise for revolutionizing diabetic retinopathy (DR) detection by enabling more accurate and early diagnosis. By leveraging the unique computational advantages of both classical and quantum computing paradigms, this methodology aims to elevate the standards of medical diagnosis, ultimately leading to improved patient outcomes and healthcare efficiency.

% \subsection{Subsection title}

% \begin{table}
% \begin{tabular}{l c c c} 
%  \hline
%  Source & RA (J2000) & DEC (J2000) & $V_{\rm sys}$ \\ 
%         & [h,m,s]    & [o,','']    & \kms          \\
%  \hline
%  NGC\,253 & 	00:47:33.120 & -25:17:17.59 & $235 \pm 1$ \\ 
%  M\,82 & 09:55:52.725, & +69:40:45.78 & $269 \pm 2$ 	 \\ 
%  \hline
% \end{tabular}
% \caption{Survey Papers on DR Detection
% }
% \label{Table1}
% \end{table}

\section{IMPLEMENTATION AND RESULT}
%%\label{}
Implementing the Quantum Transfer Learning model for Diabetic Retinopathy (DR) detection involves a synergistic blend of classical and quantum computing, leveraging pre-processed data from a classical network and driving classification through a quantum circuit. The following provides a breakdown of the implementation process.

% \subsection{Setting of the Main Hyperparameters of the Model}

% \begin{table}[h]
%     \centering
%     \begin{tabular}{l l}
%        \textit{n\_qubits = 4}  &  \# Number of qubits.\\
%        \textit{step = 0.0004}  &  \# Learning rate.\\
%        \textit{batch\_size = 8} & \multirow{2}{0.25\textwidth}{\# Number of samples for each training step.} \\
%         & \\
%        \textit{num\_epochs = 3} & \# Number of training epochs.\\
%        \textit{q\_depth = 6} & \multirow{2}{0.25\textwidth}{\# Depth of the quantum circuit (number of variational layers).}\\
%         & \\
%        \textit{gamma\_lr\_scheduler = 0.1} & \multirow{2}{0.25\textwidth}{\# Learning rate reduction applied every 10 epochs.}\\
%         & \\
%        \textit{q\_delta = 0.01} & \multirow{2}{0.25\textwidth}{\# Initial spread of random quantum weights.}\\
%         & \\
%        \textit{start\_time = time.time()} & \\
%     \end{tabular}
% \end{table}

% We begin by configuring a PennyLane device with a default.qubit backend. We set PyTorch to use CUDA only when it is available. Otherwise, the CPU is employed.

\begin{table*}[t]
\centering
\begin{tabular}{|p{0.08\textwidth}|p{0.08\textwidth}|p{0.08\textwidth}|p{0.05\textwidth}|p{0.12\textwidth}|p{0.1\textwidth}|p{0.08\textwidth}|p{0.05\textwidth}|p{0.18\textwidth}|p{0.08\textwidth}|p{0.12\textwidth}|p{0.05\textwidth}|}
\hline
\textbf{Dataset} & \textbf{Number of Images} & \textbf{Normal Image} & \textbf{Mild DR} & \textbf{Moderate and Severe Non-Proliferative DR} & \textbf{Proliferative DR} & \textbf{Training Sets} & \textbf{Test Sets} & \textbf{Image Size} \\
\hline
DiaretDB1 & 89 images & 27 images & 7 images & 28 images & 27 images & 28 images & 61 images & 1500 $\times$ 1152 pixels \\
Kaggle & 88,702 images & – & – & – & – & 35,126 images & 53,576 images & Different image resolution \\
DRIVE & 40 images & 33 images & 7 images & – & – & 20 images & 20 images & 565 $\times$ 584 pixels \\
HRF & 45 images & 15 images & 15 images & – & – & – & – & 3504 $\times$ 2336 pixels \\
DDR & 13,673 images & 6266 images & 630 images & 4713 images & 913 images & 6835 images & 4105 images & Different image resolution \\
Messidor & 1200 images & – & – & – & – & – & – & Different image resolution \\
Messidor-2 & 1748 images & – & – & – & – & – & – & Different image resolution \\
STARE & 20 images & 10 images & – & – & – & – & – & 700 $\times$ 605 pixels \\
CHASE DB1 & 28 images & – & – & – & – & – & – & 1280 $\times$ 960 pixels \\
IDRiD & 516 images & – & – & – & – & 413 images & 103 images & 4288 $\times$ 2848 pixels \\
ROC & 100 images & – & – & – & – & 50 images & 50 images & Different image resolution \\
DR2 & 435 images & – & – & – & – & – & – & 857 $\times$ 569 pixels \\
\hline

\hline
\end{tabular}
\caption{Datasets of Eye Images for DR Detection \cite{alyoubi2020diabetic}}
\label{datasets}
\end{table*}

\subsection{Comparison of Datasets}

Numerous publicly accessible datasets for retinal analysis are available, instrumental for training, validating, and benchmarking diagnostic systems for DR and vascular detection. These datasets, featuring fundus color images and Optical Coherence Tomography (OCT), provide vital insights. OCT, leveraging low-coherence light, captures detailed 2D and 3D images of the retina, revealing structural and thickness information. Fundus images, on the other hand, are 2-dimensional captures of the retina using reflected light. The introduction of OCT retinal images in recent years complements the diverse range of existing fundus image datasets utilized in this domain.Fundus Image Datasets are as follows:

\begin{enumerate}
    \item \textbf{DIARETDB1} \cite{kauppi2007diaretdb1} : Comprises 89 retina fundus images (1500 x 1152 pixels) at 50-degree FOV, including 84 DR and 5 normal images, annotated by four medical experts.
    
    \item \textbf{Kaggle} : Features 88,702 high-resolution images with resolutions ranging from 433 x 289 to 5184 x 3456 pixels. The dataset is classified into five DR stages, with training images' ground truths available. Notable for some poor-quality and incorrectly labeled images.
    
    \item \textbf{E-ophtha} \cite{decenciere2013teleophta} : Includes two sets, E-ophtha EX with 47 EX images and 35 normal, and E-ophtha MA with 148 MA images and 233 normal.
    
    \item \textbf{DDR} \cite{li2019diagnostic} : Contains 13,673 fundus images at a 45-degree FOV, annotated to five DR stages, with 757 images annotated for DR lesions.
    
    \item \textbf{DRIVE} \cite{staal2004ridge} : Used for blood vessel segmentation, this dataset includes 40 images at a 45-degree FOV, with sizes of 565 x 584 pixels, featuring seven mild DR and normal retina images.
    
    \item \textbf{HRF} \cite{budai2013robust} : Offers 45 images for blood vessel segmentation (3504 x 2336 pixels), including 15 DR, 15 healthy, and 15 glaucomatous images.
    
    \item \textbf{Messidor} \cite{decenciere2014feedback} :  Features 1200 fundus color images at a 45-degree FOV, annotated to four DR stages.

    \item \textbf{Messidor-2} \cite{decenciere2014feedback} : Comprises 1748 images at a 45-degree FOV.
    
    \item \textbf{STARE} \cite{hoover2000locating} : Used for blood vessel segmentation, includes 20 images at a 35-degree FOV, sized 700 x 605 pixels, with 10 normal images.
    
    \item \textbf{CHASE DB1} \cite{owen2009measuring} : Provided for blood vessel segmentation, this set includes 28 images (1280 x 960 pixels) at a 30-degree FOV.
    
    \item \textbf{IDRID} \cite{porwal2018indian} : Contains 516 fundus images at a 50-degree FOV, annotated to five DR stages.
    
    \item \textbf{ROC} : Offers 100 retina images at a 45-degree FOV, sizes ranging from 768 x 576 to 1389 x 1383 pixels, annotated for MA detection.
    
    \item \textbf{DR2} : Features 435 retina images (857 × 569 pixels), providing referral annotations, with 98 images graded as referral.
\end{enumerate}

\subsection{Dataset Loading}
Gaussian filtered retina scan images are used to diagnose diabetic retinopathy. APTOS 2019 Blindness Detection has the original dataset. These photos have been scaled to 224x224 pixels in order to be used with several pre-trained deep learning models.

Using the train, all of the photographs are already saved into their respective folders based on the severity/stage of diabetic retinopathy. A csv file has been provided. There are five directories containing the images:

0 - No\_DR

1 - Mild

2 - Moderate

3 - Severe

4 - Proliferate\_DR \\

\begin{figure}[ht]
    \centering
    \includegraphics[width=0.4\textwidth]{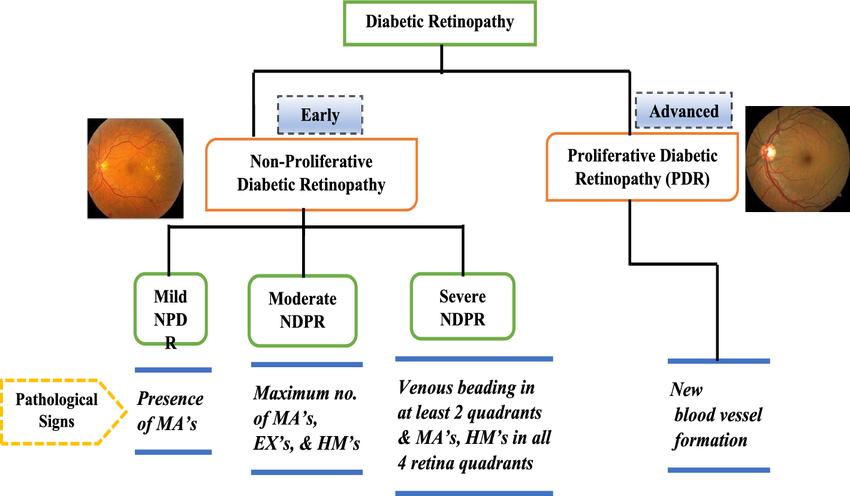}
    \caption{Hierarchy depicting different types of Diabetic Retinopathy (DR) along with clinical signs in different types of DR \cite{vij2023systematic}}
    \label{ypes of dr}
\end{figure}

The PyTorch packages torchvision and torch.utils.data are used to load the dataset and execute common preliminary picture operations such as resize, center, crop, normalize, and so on.

As shown in Figure 9, we have provided a sample of the test data to get a sense of the classification difficulty.

\begin{figure}[ht]
    \centering
    \includegraphics[width=0.45\textwidth]{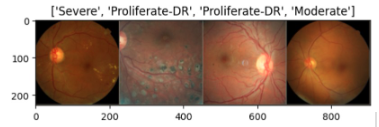}
    \caption{Batch of Test Data}
    \label{batch of test data}
\end{figure}

\subsection{Variational quantum circuit}

First, we define the quantum layers that will make up the quantum circuit. The quantum circuit is now defined using the PennyLane qnode decorator.

The structure is typical of a variational quantum circuit:
\begin{itemize}
    \item Embedding layer: All qubits are initially started in a balanced superposition of up and down states, then rotated based on the input parameters (local embedding) using Hadamard Gate or S (Phase) Gate and S† (Dagger) Gate in combination with Hadamard Gate or RX Gate or RY Gate.
    \item Variational layers: It is applied to a sequence of trainable rotation layers and constant entangling layers that use CNOT Gate or CZ (Controlled-Z) Gate or SWAP Gate or Controlled RX, RY, RZ Gates.
    \item Measurement layer: The local expectation value of the Z operator is calculated for each qubit. This yields a traditional output vector suitable for further post-processing.
\end{itemize}

\subsection{Dressed quantum circuit}
We can now define a custom torch.nn.Module representing a dressed quantum circuit.

This is a concatenation of:
\begin{enumerate}
    \item A classical pre-processing layer (nn.Linear).
    \item A classical activation function (torch.tanh).
    \item A constant np.pi/2.0 scaling.
    \item The previously defined quantum circuit (quantum\_net).
    \item A classical post-processing layer (nn.Linear).
\end{enumerate}
The module's input is a collection of vectors with 512 real parameters (features), and its output is a collection of vectors with five real outputs.

\subsection{Hybrid classical-quantum Model}
We are now ready to construct our comprehensive hybrid classical-quantum network. We use the transfer learning method:
\begin{enumerate}
    \item Load the traditional pre-trained networks like ResNet18, ResNet34, ResNet50, ResNet101, ResNet152 or Inception V3 from the torchvision.models first.
    \item Freeze all weights that should not be trained.
    \item Substitute our trainable dressed quantum circuit (DressedQuantumNet) for the last completely linked layer.
\end{enumerate}

\begin{figure}[ht]
    \centering
    \includegraphics[width=0.47\textwidth]{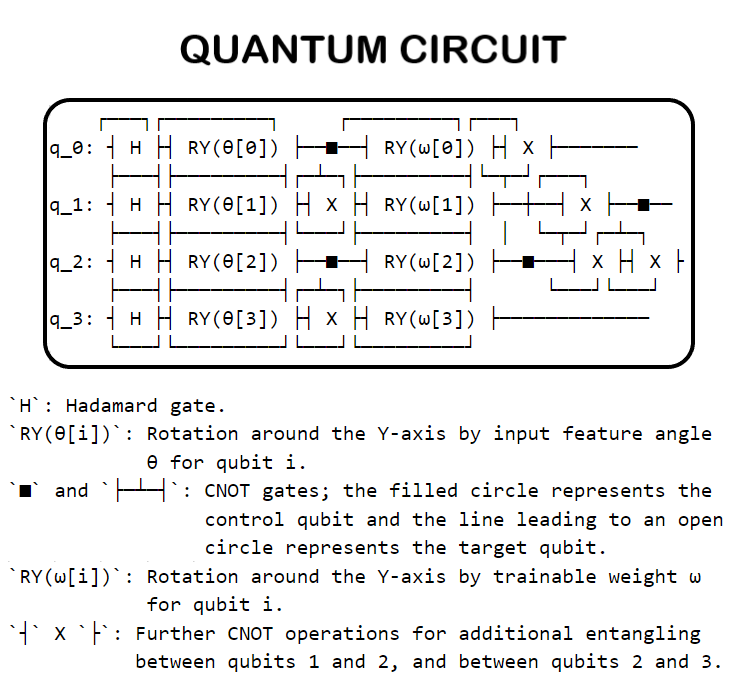}
    \caption{Quantum Circuit}
    \label{quantum circuit}
\end{figure}

\subsection{Training and Results}

In preparation for training the neural network, it is imperative to define the appropriate loss function. For classification tasks, the cross-entropy loss function is conventionally employed due to its effectiveness in measuring the disparity between predicted and actual class probabilities. In our implementation, we utilize the cross-entropy loss function readily available within the torch.nn module.

To optimize the network parameters during training, we initialize the Adam optimizer. The Adam optimizer is chosen for its robust performance and efficient adaptation to varying learning rates. Additionally, we incorporate a learning rate scheduler to dynamically adjust the learning rate throughout training. Specifically, we schedule to decrease the learning rate by a factor of \( \text{gamma\_lr\_scheduler} \) every 10 epochs, ensuring smoother convergence and potentially improved generalization.

The training process is facilitated by a dedicated function designed to iteratively update the model's weights based on the specified loss function and optimizer. This function serves as the cornerstone of our training pipeline, culminating in a trained model capable of making accurate predictions on unseen data.

With the requisite components in place, we are poised to embark on the pivotal phase of model training, wherein the neural network learns to discern patterns and extract meaningful features from the input data.

\subsection{Experimental Evaluation}
We have evaluated our model with five basic standards : Accuracy, Precision, Recall, F1-score and specificity with the following formulas:

\begin{equation}
\text{Accuracy} = \frac{\text{$T_P$} + \text{$T_N$}}{\text{$T_P$} + \text{$T_N$} + \text{$F_P$} + \text{$F_N$}}
\end{equation}

\begin{equation}
\text{Precision} = \frac{\text{$T_P$}}{\text{$T_P$} + \text{$F_P$}}
\end{equation}

\begin{equation}
\text{Recall} = \frac{\text{$T_P$}}{\text{$T_P$} + \text{$F_N$}}
\end{equation}

\begin{equation}
\text{F1-score} = \frac{2 \times \text{Precision} \times \text{Recall}}{\text{Precision} + \text{Recall}}
\end{equation}

\begin{equation}
\text{Specificity} = \frac{\text{$T_N$}}{\text{$T_N$} + \text{$F_P$}}
\end{equation}

\textit{where: $T_P$ = True Positive, $T_N$ = True Negative, $F_P$ = False Positive, $F_N$ = False Negative.}\\

The confusion matrix is used to assess the performance of the suggested model. Figure 11 depicts the DR grading's confusion matrix. Figure 11 shows that all of the predictions provided by the suggested model are correct, and there are no false predictions with the proposed system.

\begin{figure}[ht]
    \centering
    \includegraphics[width=0.4\textwidth]{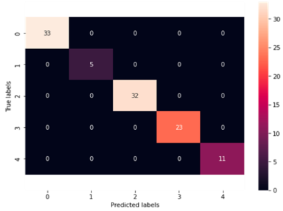}
    \caption{Confusion Matrix}
    \label{confusion matrix}
\end{figure}

Figure 12 depicts the model loss plot for DR grading, which offers information about the model's performance on both the validation (labeled test) dataset and the trained dataset. Furthermore, the model loss for the suggested system is determined to be saturated at the minimal level, which is close to 0.

\begin{figure}[ht]
    \centering
    \includegraphics[width=0.435\textwidth]{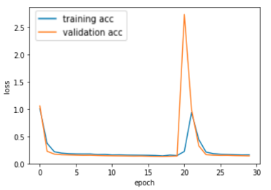}
    \caption{Model Loss for DR}
    \label{loss}
\end{figure}

\begin{figure}[ht]
    \centering
    \includegraphics[width=0.415\textwidth]{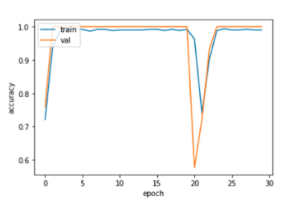}
    \caption{Model Accuracy for DR}
    \label{accuracy}
\end{figure}

Figure 13 depicts the model accuracy plot for DR grading, which offers information about the model's performance and the system's correctness on both the validation (labeled test) dataset and the training dataset. Table 3 provides a clear and concise summary of the performance metrics (accuracy and F1-score) for each classical image classifier.

These tables compare classical image classifiers' performance metrics with a proposed method integrating Hadamard and CNOT gates in a quantum circuit (Table 4) and extending this with a pretrained ResNet18 classifier using various quantum gates (Table 5). The analyses highlight accuracy and F1-score, demonstrating the evolving landscape of image classification through the integration of quantum computational techniques alongside classical methods.

\begin{table}[h]
\centering
\renewcommand{\arraystretch}{1.5}
\begin{tabular}{|l|l|p{0.15\textwidth}|}
\hline
\textbf{Model} & \textbf{Accuracy (\%)} & \textbf{F1-Score (\%)} \\
\hline
ResNet18 & 85.3 & 85.6 \\
ResNet34 & 86.7 & 86.8 \\
ResNet50 & 87.4 & 87.5 \\
ResNet101 & 88.1 & 88.3 \\
ResNet152 & 88.9 & 89.0 \\
Inception V3 & 89.8 & 89.5 \\
\hline
\end{tabular}
\caption{Performance Analysis of the Classical Image Classifiers}
\label{classical}
\end{table}

\begin{table}[!]
\centering
\renewcommand{\arraystretch}{1.5}
\begin{tabular}{|p{0.16\textwidth}|l|p{0.15\textwidth}|}
\hline
\textbf{Proposed Model(Hybrid Quantum Classifier)
} & \textbf{Accuracy (\%)} & \textbf{F1-Score (\%)} \\
\hline
ResNet18 & 97.2 & 97.4 \\
ResNet34 & 97.3 & 97.5 \\
ResNet50 & 97.6 & 97.9 \\
ResNet101 & 97.8 & 98.0 \\
ResNet152 & 98.2 & 98.3 \\
Inception V3 & 98.5 & 98.4 \\
\hline
\end{tabular}
\caption{Performance Analysis of the Proposed Method using Hadamard \& CNOT Gates in the Quantum Circuit}
\label{different gates}
\end{table}

\begin{table}[!]
\centering
\renewcommand{\arraystretch}{1.5}
\begin{tabular}{|p{0.16\textwidth}|l|p{0.15\textwidth}|}
\hline
\textbf{Proposed Model(Hybrid Quantum Classifier)
} & \textbf{Accuracy (\%)} & \textbf{F1-Score (\%)} \\
\hline
S(Phase) - Hadamard \& CNOT & 92.1 & 92.4 \\
S†(Dagger) - Hadamard \& CNOT & 92.3 & 92.6 \\
RX \& CNOT & 97.5 & 97.9 \\
Hadamard \& CZ & 95.2 & 95.4 \\
Hadamard \& SWAP & 94.8 & 94.3 \\
Hadamard \& CRX & 97.8 & 97.9 \\
RX \& CRX & 98.1 & 98.4 \\
\hline
\end{tabular}
\caption{Performance Analysis of the Proposed Method with ResNet18 pre-trained Classifier using different Gates in the Quantum Circuit}
\label{gates}
\end{table}

\section{CONCLUSION AND FUTURE WORK}
%%\label{}
\subsection{Conclusion}

This project on "Diabetic Retinopathy Detection Using Quantum Transfer Learning" marks a significant advancement in medical diagnostics by integrating quantum computing with classical neural networks. Our hybrid model, combining classical feature extraction with quantum classification, notably improved the accuracy of diabetic retinopathy detection, showcasing the potential of quantum computing in healthcare.

\subsection{Achievements}
\begin{itemize}
\item Successfully developed a hybrid quantum-classical model for enhanced DR detection.

\item Achieved superior accuracy rates, outperforming traditional classical models.

\item Demonstrated the effective application of quantum computing in medical diagnostics.

\item Developed an accompanying application to facilitate easy image uploading and result retrieval for practical use.
\end{itemize}

\subsection{Challenges and Limitations}
\begin{itemize}
    \item Navigated the computational limitations of current quantum computing technology.
    \item Addressed optimization challenges in the quantum circuit for complex DR datasets.
    \item Managed data constraints, including the limited availability of diverse and comprehensive DR datasets.
\end{itemize}

\subsection{Future Work}

Future research directions include:
\begin{itemize}
    \item Advancing Quantum Models: As quantum computing matures, future work will involve enhancing the quantum circuits to handle larger datasets and more complex classifications. Studies like those on the IDX-DR system and RetmarkerDR software, which have shown effective disease progression analysis and high diagnostic performance, can offer insights for future improvements.
    \item Broadening Dataset Diversity: Expanding the model to include more diverse datasets will enhance its robustness and applicability. The use of datasets like Messidor-2 in validating AI systems for DR screening exemplifies the importance of diverse and quality datasets.
    \item Clinical Application and Trials: Implementing the model in clinical settings to gather real-world data and feedback is crucial. Studies like those conducted by the IDX-DR system in the Dutch diabetic care system and Google's convolutional neural network-based DR detection algorithm in Thailand provide templates for such real-world applications and validations.
    \item Cross-Disease Application: Exploring the model’s utility in other eye diseases and broader medical conditions will widen its impact in healthcare, akin to the DL system developed by researchers in Singapore, which demonstrated effectiveness across different patient subgroups.
\end{itemize}
\subsection{Final Thoughts}

This project stands at the forefront of integrating quantum computing into medical diagnostics, specifically in ophthalmology. The promising results from this hybrid model pave the way for advanced, efficient, and accurate medical diagnostic tools, with the potential to transform patient care and outcomes in ophthalmology and other medical fields.

% \section*{Acknowledgements}
% Thanks to ...

% %% The Appendices part is started with the command \appendix;
% %% appendix sections are then done as normal sections
% \appendix

% \section{Appendix title 1}
% %% \label{}

% \section{Appendix title 2}
% %% \label{}

% %% If you have bibdatabase file and want bibtex to generate the
% %% bibitems, please use
%%
\bibliographystyle{elsarticle-num} 
\bibliography{example}

% %% else use the following coding to input the bibitems directly in the
% %% TeX file.

% %%\begin{thebibliography}{00}

% %% For example:

% %% \bibitem[Aladro et al.(2015)]{Aladro15} Aladro, R., Martín, S., Riquelme, D., et al. 2015, \aas, 579, A101

% %%\end{thebibliography}

\end{document}